\documentclass[conference]{IEEEtran}
\IEEEoverridecommandlockouts

\usepackage{cite}
\usepackage{amsmath,amssymb,amsfonts}
\usepackage{algorithmic}
\usepackage{graphicx}
\usepackage{textcomp}
\usepackage{listings}
\usepackage{hyperref}

\usepackage{xcolor}
\usepackage{url}
\def\BibTeX{{\rm B\kern-.05em{\sc i\kern-.025em b}\kern-.08em
    T\kern-.1667em\lower.7ex\hbox{E}\kern-.125emX}}
\begin{document}

\title{Evaluating Transformer Models for Suicide Risk Detection on Social Media\\

}

\author{
\IEEEauthorblockN{Jakub Pokrywka}
\IEEEauthorblockA{\textit{Adam Mickiewicz University} \\
Poland \\
jakub.pokrywka@amu.edu.pl}
\and
\IEEEauthorblockN{Jeremi I. Kaczmarek}
\IEEEauthorblockA{
    \textit{Adam Mickiewicz University,}\\
    \textit{Poznan University of Medical Sciences}\\
    Poland\\
    jeremi.kaczmarek@amu.edu.pl
}
\and
\IEEEauthorblockN{Edward J. Gorzelańczyk}
\IEEEauthorblockA{
    \textit{Kazimierz Wielki University,} \\
    \textit{The Society for the Substitution} \\
    \textit{"Medically Assisted Recovery"} \\
    Poland \\
    medsystem@medsystem.com.pl
}
}

\maketitle

\begin{abstract}
The detection of suicide risk in social media is a critical task with potential life-saving implications. This paper presents a study on leveraging state-of-the-art natural language processing solutions for identifying suicide risk in social media posts as a submission for the "IEEE BigData 2024 Cup: Detection of Suicide Risk on Social Media" conducted by the kubapok team. We experimented with the following configurations of transformer-based models: fine-tuned DeBERTa, GPT-4o with CoT and few-shot prompting, and fine-tuned GPT-4o. The task setup was to classify social media posts into four categories: indicator, ideation, behavior, and attempt. Our findings demonstrate that the fine-tuned GPT-4o model outperforms two other configurations, achieving high accuracy in identifying suicide risk. Notably, our model achieved second place in the competition. By demonstrating that straightforward, general-purpose models can achieve state-of-the-art results, we propose that these models, combined with minimal tuning, may have the potential to be effective solutions for automated suicide risk detection on social media.
\end{abstract}


\section{Introduction}
\subsection{Suicide and Suicidal Behavior: Definitions, Distinctions, and Determinants}

Suicide is a major global health concern, with over 720,000 individuals taking their own lives each year (WHO 2024). The true scale of the problem is likely underestimated, as many non-fatal attempts remain unreported. These behaviors often result in injury, disability, and long-lasting psychological trauma, extending beyond individuals to affect families and communities. The broad social and economic implications further highlight the urgent need for effective public health strategies\footnote{\url{https://www.who.int/news-room/fact-sheets/detail/suicide}}

While the definition of suicide is generally consistent in the literature, terminology for related concepts varies widely. Specialist psychiatric and psychological sources often use these terms inconsistently. To ensure clarity, we have adopted a simplified set of terms derived from multiple sources \cite{klonsky2016suicide} \footnote{\url{https://www.nimh.nih.gov/health/statistics/suicide}}. We use these terms to describe psychological and psychiatric phenomena. However, note that some may carry different meanings in our work when referring to the suicide risk level labels used in the competition dataset\cite{li2022suicide}. Table \ref{tab:terms} provides a comparison of these terms and their corresponding definitions.

\begin{table}[h!]
    \centering
    \caption{Comparison of the Suicide Risk Level Labels Utilized
in the Training Data with Psychiatric
and Psychological Terminology
Adapted for this Work.}
    \begin{tabular}{|p{1.1cm}|p{3.3cm}|p{3.3cm}|}
    \hline
    \textbf{Term}& \textbf{Benchmark Dataset} \cite{8665525,li2022suicide} & \textbf{This Work} \\ \hline
         Indicator&  The post content has no explicit expression concerning suicide.&  -\\
         \hline 
         Ideation &  The post content has explicit suicidal expression, but there is no plan to commit suicide.&  Thoughts, considerations, and plans of suicide.\\  \hline 
         Behavior&  The post content has explicit suicidal expression and a plan to commit suicide or self-harming behaviors.& Self-directed, injurious behavior with an intent to die.\\ \hline 
         Attempt&  The post content has explicit expressions concerning historic suicide attempts.&  Non-fatal, self-directed, injurious behavior with an intent to die.\\ \hline 
         Suicide&  -&  Fatal, self-directed, injurious behavior with an intent to die.\\ \hline
    \end{tabular}
 \label{tab:terms}
\end{table}

Suicidal ideation refers to ruminating, thinking, or planning suicide. Suicidal behavior, on the other hand, is defined as an act of self-harm performed with the intent to die. If the behavior results in death, it is classified as suicide; if it does not, it is considered a suicide attempt. Terms like “failed” or “unsuccessful” attempt should be avoided, as they imply that death is the desired outcome\cite{klonsky2016suicide}.

It is critical to differentiate suicidal behavior from non-suicidal self-injury (NSSI), which involves deliberate self-harm without lethal intent. Although NSSI and suicidal behavior may share overlapping risk factors, they serve distinct psychological purposes, necessitating precise differentiation for accurate clinical assessment and intervention\cite{nock2008suicide,klonsky2016suicide}.

The etiology of suicidal behavior is complex and multifactorial, involving psychological, social, cultural, biological and environmental components. In clinical practice, the progression from suicidal ideation to planning and attempting is often used as a framework to understand the interplay between risk and protective factors. Identifying these predictors is critical for risk assessment and adequate intervention\cite{klonsky2016suicide}.

\subsection{Motivation for Suicide Risk Detection Algorithms on the Internet}

The rapid pace of societal changes, characterized by digitalization and increased information accessibility, has introduced new dimensions to the etiology of suicide. Social media, in particular, has garnered attention as a potential contributor to the deterioration of mental health, especially among youth. Several phenomena, mainly cyberbullying and trolling, displacement of beneficial activities, and persistent preoccupation, are often regarded as exacerbating existing risk factors or acting as independent contributors, correlating with increased prevalence of self-harm, suicidal ideation, and other mental health problems\cite{khalaf2023impact}.

However, these same platforms, when combined with advancements in technology such as machine learning and big data analytics, offer promising opportunities for suicide prevention. Novel interventions leveraging social media data could reach individuals who might not otherwise receive help, using information that is typically inaccessible to mental health professionals by traditional means. This is particularly relevant given the limited access to psychiatric and psychological care and the demand for it continuously increasing. Considering that suicide is the third leading cause of death among individuals aged 15–29 (WHO, 2024) — a demographic with a high social media consumption — such solutions could be especially impactful.

One promising approach involves suicide risk detection systems deployed within social media platforms. These systems can analyze vast and diverse datasets, including the content of posts and comments, precise metadata such as time stamps and user interactions, and employ Natural Language Processing (NLP) to identify emotions, sentiments, and specific risk factors indicative of suicidal behavior. Moreover, advanced Artificial Intelligence (AI) models have the potential to discern subtle patterns suggestive of underlying somatic or neuropsychiatric symptoms. Detection systems enable continuous content monitoring, facilitate early intervention, and offer the possibility of preventing numerous tragedies through timely and targeted responses. Although detecting suicide risk is the primary objective of current systems, the ability to differentiate between genuine suicidal intent and self-harm threats without the intent to die is an intriguing perspective worth mentioning.

\subsection{kubapok at IEEE BigData 2024 Cup: Detection of Suicide Risk on Social Media}

In recent years, the field of NLP developed rapidly. This may be attributed mainly to artificial neural network architecture Transformers \cite{vaswani2017attention} and training neural models with massive amounts of text. The excellent results of the contemporary NLP models are promising for the development of suicidal detectors based on Internet texts. However, the question arises: which NLP method would work the best for this task?

The presented paper contributes to the subject of suicide risk detection on the Internet by comparing three state-of-the-art NLP approaches on a benchmark dataset developed especially for this task. The work was created as an entry for the "IEEE BigData 2024 Cup for the Detection of Suicide Risk on Social Media" \cite{li2022suicide} by the kubapok team. The benchmark dataset includes a training set and two test sets (preliminary and final). The training set comprises both annotated and non-annotated samples. Each annotated sample consists of a text with exactly one associated label: indicator, ideation, behavior, or attempt.

The approaches for such a task may include incorporating specialized NLP models designed for medical applications, data augmentation, and using pseudo-labels from non-annotated parts of the train datasets or external datasets. Moreover, the class imbalance in labels poses an additional challenge, which may be addressed by sampling techniques, adjusting class probabilities during inference, or model fine-tuning.

Nevertheless, we choose not to employ highly specialized techniques. We set our objective to utilize just a straightforward and generalizable approach by leveraging neural models of general utility without integrating advanced methods. Such an approach makes it easy to migrate it between different text domains and the availability of corpora. We evaluated two Transformer-based models: the DeBERTa model \cite{he2020deberta} (in both base and large sizes) and the GPT-4o model \cite{openai2023gpt4}. The DeBERTa model, an encoder-only architecture, is publicly available and requires fine-tuning on a task-specific training dataset to achieve optimal performance. Conversely, the GPT-4o model, a decoder-only architecture, can be accessed publicly through an API and used without fine-tuning by employing direct querying. Additionally, we explored the Chain of Thought (CoT) methodology, which prompts the model to generate intermediate reasoning steps before providing a final response, a technique shown to enhance model performance \cite{wei2022chain}. Another effective strategy is few-shot prompting, which involves providing the model with a small set of task-specific examples in the prompt \cite{brown2020language}. We incorporated both of these techniques into the prompt design for the GPT-4o model. While decoder-only models can also be fine-tuned to improve accuracy, this capability for the GPT-4o model has only recently become available through its API. We experimented with fine-tuning approach as well.

To sum up, we tested the following methods:
\begin{itemize}
    \item DeBERTa (encoder-only model) for classification task 
    \item GPT-4o (decoder-only model) with few-short prompting and Chain of Thought (CoT)
    \item GPT-4o (decoder-only model) fine-tuned generative model
\end{itemize}

Both model families were evaluated in the context of the SemEval 2024 Task 9 Brainteaser challenge \cite{li-etal-2024-hw}, a recent competition focused on a classification task outside the medical domain. In this paper, we demonstrate that among the methods we evaluated, the GPT-4o model with fine-tuning achieved the best performance. Our final solution, based on this fine-tuned GPT-4o, scored second place in the "IEEE BigData 2024 Cup for the Detection of Suicide Risk on Social Media", finishing marginally behind the top team among the 13 teams that participated in the final evaluation. This outcome illustrates that even straightforward, general-purpose models can achieve state-of-the-art performance, provided they leverage advanced model architectures effectively.

\section{Related Work}

The use of NLP in suicide risk detection has evolved significantly. Initial research focused on traditional machine learning methods, such as Support Vector machines (SVMs) and logistic regression. For example, a study\cite{ws-2008-current} demonstrated that these methods could distinguish between genuine and elicited suicide notes with accuracy comparable to mental health professionals, highlighting the potential of computational tools in assessing suicide risk.

As social media platforms became more widely used, research shifted to analyzing user-generated content to detect signs of suicidal ideation. For example, a study\cite{aladaug2018detecting} employed text mining techniques, including Term Frequency-Inverse Document Frequency (TF-IDF) and sentiment analysis, which were applied to posts from Reddit forums like r/SuicideWatch, r/Depression, and r/Anxiety. The study utilized machine learning models like Logistic Regression, SVMs, and Random Forest (RF), achieving high accuracy (80\%-92\%) in distinguishing suicidal posts from non-suicidal ones, demonstrating the effectiveness of these methods for identifying at-risk individuals online.

The introduction of original Transformer architecture\cite{vaswani2017attention} and its variants, like BERT\cite{devlin-etal-2019-bert} and GPT\cite{radford2018improving}, has revolutionized NLP. Such models, built on the self-attention mechanism, have proven to be especially effective in capturing nuanced language patterns and contextual dependencies. Their ability to model complex semantics has made them particularly suitable for mental health applications, including suicide risk detection.

Recent research has shown that Transformer-based approaches outperform traditional machine-learning methods in detecting suicidal ideation. One study\cite{long2022quantitative} evaluated various models on social media datasets comprising Twitter and Reddit posts. The results demonstrated that Transformer models, particularly BERT and RoBERTa\cite{liu2019robertarobustlyoptimizedbert}, achieved significantly higher accuracy than traditional approaches like SVMs and RF. The advantage of these models was most notable when analyzing longer texts from Reddit, where they demonstrated greater accuracy and contextual understanding. In 2019, a shared task was conducted to detect the degree of suicide risk in Reddit posts \cite{zirikly-etal-2019-clpsych}. The authors of \cite{mohammadi2019clac} achieved first place in two out of the three subtasks using SVMs. Two teams utilized BERT for the competition \cite{ambalavanan2019using, matero2019suicide}. Aside from the competition, other researchers have utilized neural models different from transformer-based architectures — specifically, Long Short-Term Memory (LSTM) and Convolutional Neural Networks (CNN) — for the Bangla language, as advanced transformer-based NLP models were not yet available at that time \cite{GHOSH2023119007}. Authors of \cite{mirtaheri2024self} used custom self-attention LSTM-based custom architecture enriched by CNN and BERT embeddings.


In recent studies, the use of transformer-based models has become increasingly common. For instance, the authors of \cite{haque2020transformer} and \cite{metzler2022detecting} conducted comparisons between traditional BiLSTM and TF-IDF-based models with transformer models and reported that transformer-based approaches significantly outperformed the conventional methods for suicidal ideation detection. Similarly, other studies have also leveraged transformer models for this task, such as those presented in \cite{boonyarat2024leveraging, zhang2024ketch, ghanadian2024socially, kodati2024emotion}. Authors of \cite{inan2023llamaguardllmbasedinputoutput}
developed a Large Language Model (LLM)--based safeguard model for safety risk detection in other LLM prompts. Among many labels, there is the Suicide \& Self Harm category.

\section{Suicide Detection Task}

\subsection{Original Dataset}

Reference \cite{li2022suicide} developed a dataset, primarily utilizing Reddit’s r/SuicideWatch subreddit, collecting posts from January 2020 to December 2021. The initial dataset consisted of 139,455 posts from 76,186 users. Preprocessing involved the removal of personally identifiable information and the elimination of duplicates. Posts were filtered using negative expressions and terms prominently associated with suicidal ideation in order to identify users exhibiting potential suicide risk.

These users’ historical posts and comments from 14 relevant subreddits (e.g., r/Depression, r/SelfHarm) were then gathered. The last post of each user was designated as the “targeted post” to serve as a representation of their latest suicidal risk. This process resulted in a final dataset of 3,998 posts from 1,791 users. To provide additional contextual information, posts made by a user in r/SuicideWatch or in one of the 14 relevant subreddits in the week prior to the targeted post, along with their and other users’ comments made under the targeted post, were further included in the dataset.

Manual annotation was conducted on a subset of 500 users, with posts categorized into four suicide risk levels: indicator, ideation, behavior, and attempt. This annotation was carried out at two levels: based solely on the post and with consideration of contextual information. Additionally, at the sentence level, labels for 17 categories of suicide triggers were added. These triggers were specific events or experiences, such as mental disorders, helplessness, or substance use, and were regarded by the authors as potential catalysts that could push someone closer to attempting suicide.

It is important to note that the training data was narrowed in the competition and did not include trigger labels or contextual information. We further describe the dataset that was made available to us to complete the task, which derived from the previous work.

\subsection{Competition Dataset}

This subsection provides an overview of the task titled "IEEE BigData 2024: Detection of Suicide Risk on Social Media." The training dataset, derived from prior work \cite{li2022suicide}, consisted of 2,000 Reddit posts, of which 500 were annotated and 1,500 remained unannotated. The preliminary test set contained 100 samples with hidden labels, which were used to assess the teams' initial results during the competition. Following the competition, all submitted models were evaluated using a separate test set for the final assessment. Each annotated sample in the dataset was categorized into one of four classes of suicide risk level: indicator, ideation, behavior, and attempt. The official evaluation metric for this shared task was the weighted F1-score (wF1). Tables \ref{tab:post_risk_all} and \ref{tab:post_risk_training} provide a detailed breakdown of the complete dataset and the training subset, respectively.

The dataset exhibited a relatively balanced representation of posts across the indicator, ideation, and behavior categories. In contrast, the attempt category was significantly underrepresented, creating a challenge due to the imbalance in category distribution. Additionally, posts in the attempt category had a higher average character and word count compared to the other categories. This suggests that distinguishing the attempt category from the others may be relatively straightforward, while differentiating between the remaining categories could present more difficulty.

\begin{table}[h!]
    \centering
    \caption{Distribution of Posts Across Dataset Splits}
    \begin{tabular}{|l|r|}
    \hline
        \textbf{Dataset Split} & \textbf{Number of Posts} \\
        \hline
        Training (Annotated) & 500 \\
        Training (Unannotated) & 1500 \\
        Preliminary Test & 100 \\
        \hline
    \end{tabular}
    \label{tab:post_risk_all}
\end{table}

\begin{table}[h!]
    \centering
    \caption{Distribution and Characteristics of Risk Categories in the Annotated Training Set}
    \begin{tabular}{|l|r|r|r|}
    \hline
        \textbf{Category} & \textbf{Number of Posts}  & \textbf{Avg Characters} & \textbf{Avg Words} \\
        \hline
        Indicator & 129 & 697 & 136 \\
        Ideation & 190 & 835 & 162 \\
        Behavior & 140 & 913 & 178\\
        Attempt & 41 & 1776 & 339 \\
        \hline
        \textbf{All} & 500 & 899 & 174  \\
        \hline
    \end{tabular}
    \label{tab:post_risk_training}
\end{table}

\lstset{ 
    basicstyle=\ttfamily\small,     
    breaklines=true,                
    breakindent=0pt 
}

Below, we provide one sample of each category from the training dataset. If the sample exceeds 500 characters, it is truncated. Please note that the comments may include distressing content or contain offensive language.
\\ 

Indicator:
\begin{lstlisting}
Friend may kill himself soon. Unsure what to do. I have a good friend that I met online 14 years ago, and we finally met in person last year and really connected on a much deeper level. I have been suicidal on-and-off for much of my life due to having a bad case of Crohn's Disease and because of stress, anxiety, and depression. Fortunately, I am in a good place now and haven't had suicidal ideation in months (which is a first for me). I did have one very serious attempt in 2009 which almost...
\end{lstlisting}

Ideation:
\begin{lstlisting}
Nsfw. So uh. ever since I was younger I've had a bad porn issue, one of those kids who had like a ton of porn downloaded on their phone, etc etc. I dislike it to say the least, even to this day, I dislike the roles it plays in my day to day life and find myself being disappointed in myself. That issue is something besides what i'm talking about now but it helps set the idea. My girlfriend has cheated on me multiple times and for some reason I'm still with her. I don't wanna think or talk about... 
\end{lstlisting}

Behavior:
\begin{lstlisting}
i'm gonna overdose on iron pills. i have pain everyday and i don't wanna deal with it. i'm sorry. i may do it tonight.
\end{lstlisting}

Attempt
\begin{lstlisting}
"There is nothing holding me here.so now what?. I have been wanting to off myself since 7. I am now 30.\n\nLiving situation is shit. My father, landlady and roommate's ex all say I am a problem and they all gaslight me and are narcissists. I havent been living in that house for 3 weeks in April so far. \n\nPartner/boyfriend/friend whatever he is/was stopped talking to me because of all the drama . Roommate's ex makes fake instagram accounts and messages me very harrassing fucked up things...
\end{lstlisting}

\section{Methods}

In this section, we describe our approach, which focuses on evaluating simple, universal, state-of-the-art methods. The first subsection details the use of publicly available DeBERTa models, in both base and large sizes, fine-tuned on the provided training dataset.

The subsequent subsections address the GPT-4o model, which was tested in two configurations. Subsection \ref{gptft} outlines the use of the model in an in-context learning setup, where the model’s weights remain unchanged, and all task-specific information is provided solely through the input prompt. Subsection \ref{gptnoft} describes a configuration in which the model undergoes fine-tuning. Although the GPT-4o model is not available for direct download and can only be accessed through an API, recently, it was made possible to fine-tune the model via the API, a method we adopted for our experiments.

\subsection{DeBERTa Models}

We employed two DeBERTa models —DeBERTa-base and DeBERTa-large — which we fine-tuned locally. All training and inference tasks were conducted on a single A100 GPU with 80GB VRAM. The fine-tuning process utilized the Hugging Face Transformers library \cite{wolf-etal-2020-transformers} in conjunction with the Hugging Face Trainer. Text inputs were truncated to 512 tokens, and the models were trained with the following hyperparameters: learning\_rate= 2e-5, batch\_size = 32, warmup\_ratio = 0.2, and for 100 epochs. The best-performing model was selected based on the evaluation of the weighted F1 score on a validation dataset. A random 20\% split of the data, without stratification, was reserved for testing as a validation dataset.

We explored both single-model and ensemble-model configurations. Multiple ensembles, based on different random train/validation splits, were evaluated using the competition leaderboard. The optimal ensemble was determined based on its performance on the preliminary test set provided by the challenge organizers.

\subsection{GPT-4o Without Fine-Tuning}
\label{gptnoft}
GPT-4o is currently among the top-performing Transformer models. In our approach, we leveraged the GPT-4o version, employing the few-shot learning technique. Additionally, we integrated the Chain-of-Thought method, wherein the model first generated an explanation before producing the final answer. We set the temperature parameter to 0.0 for inference to ensure deterministic outputs and the highest probability for the correct answer.

The following instruction prompt was used:

\begin{lstlisting}
You will be given social media texts (documents). Each document will be associated with one of the suicide risk labels: 'A. indicator', 'B. ideation', 'C. behavior', 'D. attempt'. Your task is to label new documents with one of these risk labels. Output a JSON with the following format: {'explanation':'', 'label':''}.
\end{lstlisting}

We evaluated two configurations of the few-shot prompting approach using 100 examples (referred to as "GPT-4o CoT 100 shot" in Table \ref{tab1} and 500 examples (referred to as "GPT-4o CoT 500 shot"). Both configurations were tested using the gpt-4o-2024-05-13 model. We only conducted evaluation on a single model and did not assess ensemble models in this case.

\subsection{GPT-4o Fine-Tunned}
\label{gptft}
Recently, the OpenAI API enabled fine-tuning of the gpt-4o-2024-08-06 model on custom data. We leveraged this capability and used the following prompt:
\begin{lstlisting}
You will be given a social media text. There are four possible suicidal risk labels: 'A. indicator', 'B. ideation', 'C. behavior', 'D. attempt'. Your task is to label the given document with one of these risk labels. e.g., 'C. behavior'.
\end{lstlisting}

We used the same prompt for inference, with a temperature of 0.0. For ensemble learning, we used a stratified 3 train/validation split (from scikit-learn's library \cite{pedregosa2011scikit} StratifiedShuffleSplit) with a test size of 10\%, training for six epochs across different seeds.

\section{Results}

\begin{table}[htbp]
\caption{Preliminary Test Scores Provided by the Competition Organizers of our Methods}
\begin{center}
\begin{tabular}{|l|r|}
\hline
\textbf{Solution} & \textbf{wF1}  \\
\hline
DeBERTa-base single model & 64.8 \\
DeBERTa-base ensemble & 69.0 \\
DeBERTa-large best ensemble & 73.0 \\
GPT-4o CoT 100 shot & 58.9 \\
GPT-4o CoT 500 shot & 58.9 \\
GPT-4o ft single model & 73.6 \\
GPT-4o ft ensemble & \textbf{74.8} \\
\hline
\end{tabular}
\label{tab1}
\end{center}
\end{table}

\begin{table}[h!]
    \centering
        \caption{Final Scores of all the Teams Participating in the Final Evaluation}
    \begin{tabular}{|r|l|r|}
    \hline
        \textbf{Rank} & \textbf{Team Name} & \textbf{wF1} \\
        \hline
        1 & Detection of Suicide & 76.1 \\
        2 & kubapok & 75.5 \\
        3 & mukumuku & 75.1 \\
        4 & BioNLP@WCM & 74.6 \\
        5 & Calculators & 73.4 \\
        6 & The Dual & 73.1 \\
        7 & BNU AI and Mental Health & 71.1 \\
        8 & MindFlow & 70.7 \\
        9 & EEEAT & 69.9 \\
        10 & MIDAS & 69.8 \\
        11 & PotatoTomato & 69.1 \\
        12 & LifeWatcher & 55.3 \\
        13 & Data Science and Decision Making... & 55.0 \\
        \hline
    \end{tabular}
    \label{tab:final_scores_percentage}
\end{table}

The results reported as weighted F1 scores on the preliminary test set provided by the competition organizers are in Table \ref{tab1}. The best-performing solution is the fine-tuned GPT-4o model. Not very far behind is the DeBERTa-large model, which was much better than the DeBERTa-base model. However, when participating in the competition, we submitted multiple solutions using many ensembles of DeBERTa models, and the results varied greatly. This may be due to the fact that the test dataset was small (100 samples) or due to the inconsistency of the DeBERTa model performance. However, for fine-tuned GPT-4o, we used only three models, and all the results were consistently high. That's why we ultimately chose fine-tuned GPT-4o for the final solution. The least performing method was GPT-4o without fine-tuning, but only with CoT and few-shot learning methods. This model performed almost the same when 100 and 500-shot examples from the training dataset were used as examples in the prompt.

Table \ref{tab:final_scores_percentage} presents the final test dataset scores of all teams participating in the competition. Our team achieved second place with a score of 75.5, closely following the top score of 76.1. According to the competition summary provided on the organizers' webpage\footnote{\url{https://competitionpolyu.github.io/report.html}}, there were 47 registered teams, of which 21 actively participated in the leaderboard. Participants employed various strategies to leverage the unlabeled training dataset, such as pseudo-labeling, manual annotation, and data augmentation. Additionally, teams addressed class imbalance using techniques like sampling strategies, class weight adjustment, and custom loss function designs. Although we experimented with class weight adjustments — ultimately finding them ineffective in our case — we did not implement other methods for handling the label imbalance.

Our approach incorporated an ensemble of the best-performing models, a widely adopted strategy in competitive settings that typically enhances the final score. Despite relying on straightforward methods, our use of state-of-the-art models proved highly effective. This outcome suggests that, in this context, established universal techniques can achieve exceptional results.

It's important to note that models like GPT4-o, which are only accessible via API, have a significant disadvantage in that it is not possible to conduct in-depth research on these models, and using them on a large scale can be expensive. Nonetheless, one key advantage is that using these models doesn't require access to GPU infrastructure for training or inference, nor does it require advanced programming skills. This ease of use makes these models accessible to a wide range of users, including medical staff without expertise in computer science.

\section{Conclusion}

In this study, we explored three approaches for the "IEEE BigData 2024 Cup: Detection of Suicide Risk on Social Media" shared task. These approaches included utilizing an encoder-only model and two configurations of a decoder-only model: one using CoT with in-context Learning and the other incorporating fine-tuning. Our results indicated that the best-performing method was the fine-tuned decoder-only model, likely due to the inherent strength of the GPT-4o. Notably, fine-tuning yielded superior performance for this particular model compared to the in-context learning setup.

Our final solution did not employ any advanced or specialized techniques beyond model ensembling. Nevertheless, despite the simplicity of our approach, it achieved second place in the competition, demonstrating the effectiveness of leveraging robust, pre-trained models and straightforward methodologies.

\section{Limitations}

Despite the strong performance of models, particularly the fine-tuned GPT-4o, when considering their use in suicide risk detection, it is essential to recognize the limitations of the dataset used in their training. In clinical practice, suicide risk assessments are conducted during a face-to-face interview by trained professionals who consider both risk and protective factors within a comprehensive evaluation of the patient’s medical history\cite{favril2022risk,park2020suicide}. While the original dataset aims to capture these elements through detailed sentence annotations and prior user posts, the competition dataset has been narrowed, containing only four broad risk categories per post and lacking contextual data. Although these categories cover key aspects such as suicidal ideation, behaviors, and past suicide attempts, the exclusion of labels for specific suicide triggers diminishes the competition dataset’s predictive value, as these labels provide critical insights into accompanying risk factors.

While representing an advancement over previous resources, the dataset is subject to certain methodological biases that may affect its generalizability. The primary data were collected from r/SuicideWatch posts between January 1, 2021, and December 31, 2022, introducing two key limitations. First, the dataset may not be representative of the broader Reddit community or users who do not engage with r/SuicideWatch. Second, the posts were collected during the COVID-19 pandemic, which may limit the dataset’s applicability under different circumstances as pandemic-related mental health effects diminish.

The creators expanded the dataset by including posts from 14 additional subreddits made by the same users, thereby increasing its scope beyond a single online community. However, this approach may still overlook crucial posts that could enhance the accuracy of suicide risk prediction. Incorporating data from a broader range of subreddits could better capture the diversity of user experiences and enhance the dataset’s representativeness.

In the original dataset, each user’s most recent post made in one of the selected subreddits was designated as the “targeted post,” with contextual data collected from comments under it and from the user’s activity in one of 14 relevant communities in the preceding week. While this contextual information is absent from the competition dataset, which contains only the "targeted post", we endorse the inclusion of previous posts in the original dataset, as it offers critical insights into the development of suicidal behavior. Nevertheless, limiting the context to one week makes it insufficient for capturing long-term risk patterns, as suicidal ideation can persist or fluctuate over extended periods\cite{nock2008suicide,klonsky2016suicide}. Encouragingly, the dataset creators have expressed a willingness to extend this timeframe.

The original dataset, which includes contextual information from previous posts and utilizes more precise labels to capture specific risk factors, represents a significant advancement in training data for models aimed at detecting suicidal behavior based on online activity. Despite only being able to use the narrow competition training set, we identified several limitations that are inherent in any dataset constructed solely from online content.

\section{Future Work}

We propose developing a dataset based on the online activity of patients whose suicidal tendencies have been confirmed or ruled out by qualified professionals, such as psychiatrists or clinical psychologists. This approach could potentially offer a more objective and reliable method for assessing suicide risk, though it is not without drawbacks.

Online posts and comments often lack verification in terms of authenticity when compared to clinical assessments, where more objective mental state examinations can be conducted. Consequently, information regarding suicidal status gathered solely from social media activity is not as reliable as the one deriving from direct clinical interviews. Although the significance of this potential problem in current datasets is unknown, thorough clinical assessment could reduce the likelihood of including content suggesting suicide risk authored by individuals whose actual risk is low or negligible. This should allow the new models to effectively distinguish whether the textual information containing claims of suicidal thoughts and behavior was made by an individual with actual desire and intention to end their own life. If this feature can be achieved without decreasing the sensitivity of the detection system, it would be most desirable as a means for reasonable resource allocation. There is a phenomenon known as "parasuicide", and suicide threats made without suicidal intentions are a common occurrence \cite{welch2001review}. With clinical assessment, the risk status would be validated more rigorously compared to the traditional annotation of posts and comments, 

The value of many predictors, such as prior suicide attempts, their frequency and lethality, acute psychological distress, the presence of mental or physical disorders, access to lethal means, and others, is generally well established \cite{favril2022risk,park2020suicide,klonsky2016suicide}. The crucial point is that many of these factors are considered predictive when they co-occur, with only several considered highly predictive on their own. In clinical practice, the presence of many factors can be established by simply asking more follow-up questions. The data not only reflects online activity but can also encompass medical information such as the onset, severity of symptoms, and history of suicide attempts, thereby providing a more comprehensive profile of individual risk factors. This does not hold for social media, where users typically disclose information they find personally significant or respond to comments that are rarely made by mental health professionals.

Moreover, this approach could enable the development of models capable of detecting subtle patterns that may elude human annotators. Not all individuals experiencing suicidal ideation or attempts explicitly communicate them online. However, their overall activity – although seemingly unrelated to suicide behavior – may still contain indicators that can be leveraged for risk prediction.

A significant challenge with this proposed approach is the inclusion of populations of individuals who have died by suicide and those at risk who do not engage with mental health services – both of whom are critical for achieving a representative dataset. Furthermore, implementing this strategy would require collaboration between data scientists and healthcare institutions, patient consent, willingness to share relevant information (e.g., social media usernames), and approval from the appropriate bioethical committee. Additionally, low patient participation could limit the feasibility of such a dataset.

A meta-analysis of 71 studies concluded that using suicidal ideation as a test for later suicide is limited \cite{mchugh2019association}. This can suggest that the absence of reported suicidal thoughts is not a definitive indicator of safety. It is noteworthy that the included studies relied on ideations reported by medical personnel, which the authors acknowledged may have underestimated cases where suicidal thoughts emerged only shortly before the suicide. It is plausible that online anonymity may encourage individuals to disclose suicidal thoughts and behaviors more openly than they would in clinical settings, potentially making online expressions of suicidal ideation a more valuable and dynamic marker compared to those obtained during clinical interviews.

It remains uncertain whether a dataset developed to recruit patients would turn out better than the existing resources. Nevertheless, even if it proves less effective, such findings would have significant implications for future research and the development of new datasets. Our goal is to collaborate with healthcare institutions and design a research protocol that adheres to ethical standards, ensuring the highest possible degree of anonymity and safety for participants. To the best of our knowledge, no such initiative has been undertaken to date. 

\section*{Acknowledgment}
We want to express our gratitude to Ryszard Staruch for the discussion and for preparing scripts to use non-annotated labels. Unfortunately, we were unable to implement this approach in time. 

We also extend our thanks to Piotr Jabłoński for suggesting some points to include in the Related Work section.

\bibliographystyle{IEEEtran}
\bibliography{biblio}

\end{document}